\documentclass[letterpaper]{article} %
\usepackage{aaai2026}  %
\usepackage{times}  %
\usepackage{helvet}  %
\usepackage{courier}  %
\usepackage[hyphens]{url}  %
\usepackage{graphicx} %
\usepackage{natbib}  %
\usepackage{caption} %
\usepackage{algorithm}
\usepackage{algorithmic}

\usepackage{tablefootnote}
\usepackage{multirow}
\usepackage{booktabs}
\usepackage{bm}
\usepackage{subfigure}
\usepackage[inline]{enumitem} 
\usepackage{xcolor}  
\usepackage{makecell}

\usepackage{pifont}

\usepackage{newfloat}
\usepackage{listings}
\DeclareCaptionStyle{ruled}{labelfont=normalfont,labelsep=colon,strut=off} %
\floatstyle{ruled}
\newfloat{listing}{tb}{lst}{}
\floatname{listing}{Listing}
\title{Robust Detection of Synthetic Tabular Data under Schema Variability}
\author{
    G. Charbel N. Kindji\textsuperscript{\rm 1,\rm 2},\\
    Elisa Fromont\textsuperscript{\rm 2},\\
    Lina M. Rojas-Barahona\textsuperscript{\rm 1},
    Tanguy Urvoy\textsuperscript{\rm 1}
}

\affiliations{
    \textsuperscript{\rm 1}Orange Labs Lannion\\
    charbel.kindji.orange@gmail.com, \{charbel.kindji, lina.rojas, tanguy.urvoy\}@orange.com

    \textsuperscript{\rm 2}Université de Rennes, CNRS, Inria, IRISA UMR 6074\\
    elisa.fromont@irisa.fr
}

\begin{document}

\maketitle

\begin{abstract}
The rise of powerful generative models has sparked concerns over data authenticity. While detection methods have been extensively developed for images and text, the case of tabular data, despite its ubiquity, has been largely overlooked. Yet, detecting synthetic tabular data is especially challenging due to its heterogeneous structure and unseen formats at test time.
We address the underexplored task of detecting synthetic tabular data ``in the wild'', \textit{i.e.} when the detector is deployed on tables with variable and previously unseen schemas. We introduce a novel datum-wise transformer architecture that significantly outperforms the only previously published baseline, improving both AUC and accuracy by 7 points. By incorporating a table-adaptation component, our model gains an additional 7 accuracy points, demonstrating enhanced robustness. This work provides the first strong evidence that detecting synthetic tabular data in real-world conditions is feasible, and demonstrates substantial improvements over previous approaches. Following acceptance of the paper, we are finalizing the administrative and licensing procedures necessary for releasing the source code. This extended version will be updated as soon as the release is complete.

\end{abstract}


\section{Introduction}
\label{section:intro}

In recent years, deep learning-based generative models have surged in popularity \cite{Suzuki19032022,RegenwetterDGM22}, raising significant concerns about their potential misuse \cite{marchal2024generativeaimisusetaxonomy}, including opinion manipulation, fraud, and harassment. In response, numerous detection methods have been developed for uniformly structured media such as images and text \cite{LiuDetectImages2022,zhu-etal-2023-beat}. However, detecting synthetic tabular data remains largely underexplored, despite the modality's importance in high-stakes domains where data integrity is crucial. It also presents unique challenges, such as heterogeneous structures, diverse feature types and distribution, and variable table sizes.
Additionally, an effective synthetic tabular data detector must be \textit{table-agnostic}, meaning it should function independently of a fixed table structure. This requirement disqualifies most state-of-the-art tabular predictors, including \cite{breiman2001random,chen2016xgboost,prokhorenkova2018catboost}, as well as recent transformer-based models tailored to specific tabular structures \cite{arik2021tabnet,somepalli2022saint}.

\cite{KindjiIDA2025} has categorized the detection of synthetic tabular data into three levels of ``wildness'':
\begin{enumerate*}[label=(\roman*)]
    \item \textit{Same-table} detection: Identifying synthetic data within a single table structure, as for the \textit{Classifier Two-Sample Test} (C2ST)~\cite{lopez2016revisiting}. In this setup, the detector does not need to be table-agnostic.
    \item \textit{Cross-table} detection: Identifying synthetic data across multiple tables (e.g. training and testing both on \textit{Adult} and \textit{Insurance} tables). This setup requires a table-agnostic detector that generalize across different tables within a predefined corpus.
    \item \textit{Cross-table shift} detection: Handling deployment scenarios where the tables encountered at inference differ from those seen during training (e.g., training across both \textit{Adult} and \textit{Insurance}, and evaluating across both \textit{Higgs} and \textit{Abalone}).
\end{enumerate*}
Each level, while independent from the others, presents an increasing challenge, and our goal in this paper is to address the most challenging one: the \emph{cross-table shift}.  In this setting, detecting synthetic versus real rows is framed as a binary classification problem. Our contributions are as follows.
\begin{itemize}
    \item We introduce a novel transformer-based architecture that is both table-agnostic and invariant to column permutations, addressing the critical need for robustness in real-world deployment.
    \item We explore a new type of distribution shift: the cross-table shift with a domain component (e.g. \textit{science} vs. \textit{finance} domain tables). This involves handling tables that differ between training and inference, both in terms of structure and domains, adding complexity to the detection task. We refer to this as \textit{cross-domain table shift}.
    \item We incorporate a \textit{table adaptation} strategy~\cite{DA2010}, resulting in notable performance improvements over existing baselines.  
\end{itemize}
Our architecture, being table-agnostic and invariant to column permutations, has the potential to be applied to tasks beyond synthetic data detection, such as regression and classification on tabular data. 
We present the related work in Section~\ref{section:related_work}, followed by the detailed description of our model in Section~\ref{section:table_agnostic_eq_transf}. We then present our experimental setup, results and limitations in Sections~\ref{section:experiments} and \ref{section:results}. Finally, we conclude and outline future research directions in Section~\ref{section:conclusion}.

\section{Related Work}
\label{section:related_work}

\begin{figure*}[h!]
    \centering
    \includegraphics[width=\textwidth]{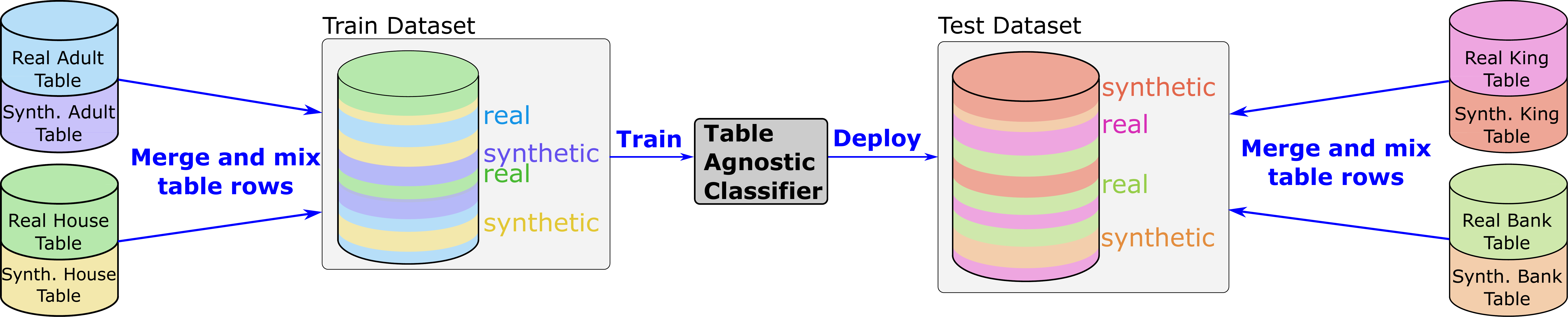}
    \caption{\textit{Cross-table shift} protocol: the real-vs-synthetic detector is trained on a mixture of table rows and tested/deployed on a mixture from holdout tables.}
    \label{fig:crosstable}
\end{figure*}

Recent research on tabular data has increasingly shifted toward the development of foundation models~\cite{kimcarte,iida-etal-2021-tabbie,herzig-etal-2020-tapas,liu2022tapex}, inspired by the impressive progress in text and vision. These models aim to produce table-agnostic representations through pretraining on diverse tables, with the goal of generalizing across tasks and schemas.
However, most of them break this table-agnosticism at inference: they rely on fixed schemas for finetuning and deployment, making them unsuitable for settings where training and test rows come from different tables. In this work, we adopt a cross-table setup (Figure~\ref{fig:crosstable}) where no schema alignment or retraining is possible, and generalization across heterogeneous table structures is required. In such settings, the main challenge is not the classifier itself, standard models could suffice, but rather the design of robust and transferable representations. As such, the following related work will primarily focus on approaches to tabular representation learning under schema variability.

Many models achieve table-agnostic pretraining via text-based encodings. TaBERT~\cite{yin20acl} and TABBIE~\cite{iida-etal-2021-tabbie} rely on the seminal BERT LLM to encode tables. TAPAS~\cite{herzig-etal-2020-tapas} and TAPEX~\cite{liu2022tapex} adapt this approach for question answering. TabuLa-8B~\cite{gardner2024large} converts table rows to text for LLM finetuning, while STab~\cite{hajiramezanali2022stab} and STUNT~\cite{nam2023stunt} emphasize generalization through data augmentation and meta-learning. Others like Xtab~\cite{XtabZhu2023}, UniTabE~\cite{yang2024unitabe}, TransTab~\cite{wang2022transtab}, PORTAL~\cite{spinaci2024portal}, and CARTE~\cite{kimcarte} instead rely on \textit{type-specific} encoders and require a fixed schema across training and testing. To tackle this problem, we introduce the \textit{Datum-wise Transformer}, a lightweight transformer trained on individual rows. Each featured is textualized and treated separately through a first transformer block to obtain an embedding for each column, then, a second transformer block without positional embedding is used on these  embeddings, enabling the model to achieve column permutation invariance and flexibility across different table structures. The model is designed for row-level training and inference on any table, with no assumption of schema consistency.

Our approach contrasts with traditional BERT-like tabular encoders such as \textit{Flat Text}~\cite{KindjiIDA2025} and TaBERT, which apply positional encodings across entire rows and thus remain sensitive to column order. While models like PORTAL, TransTab, and UniTabE also implement a form of independent feature encoding, they typically rely on partially handcrafted strategies, often conditioned on feature types or metadata. In contrast, our method learns feature representations directly from raw input without requiring data-type-specific mechanisms. %
Additionally, many BERT-based tabular models use $768$-dimensional embeddings inherited from the original architecture. In contrast, our $192$-dimensional representation aligns with our lightweight Transformer design, reducing computational and memory costs. This approach offers the advantage of building a simpler model, which has the potential to scale more easily to the large datasets commonly found in industrial contexts. This demonstrates that robust feature learning and permutation handling can compensate for reduced capacity, challenging the need for high-dimensional embeddings.

\section{Table-agnostic Datum-wise Transformer}
\label{section:table_agnostic_eq_transf}

\begin{figure*}[]
    \centering
    \includegraphics[width=\textwidth]{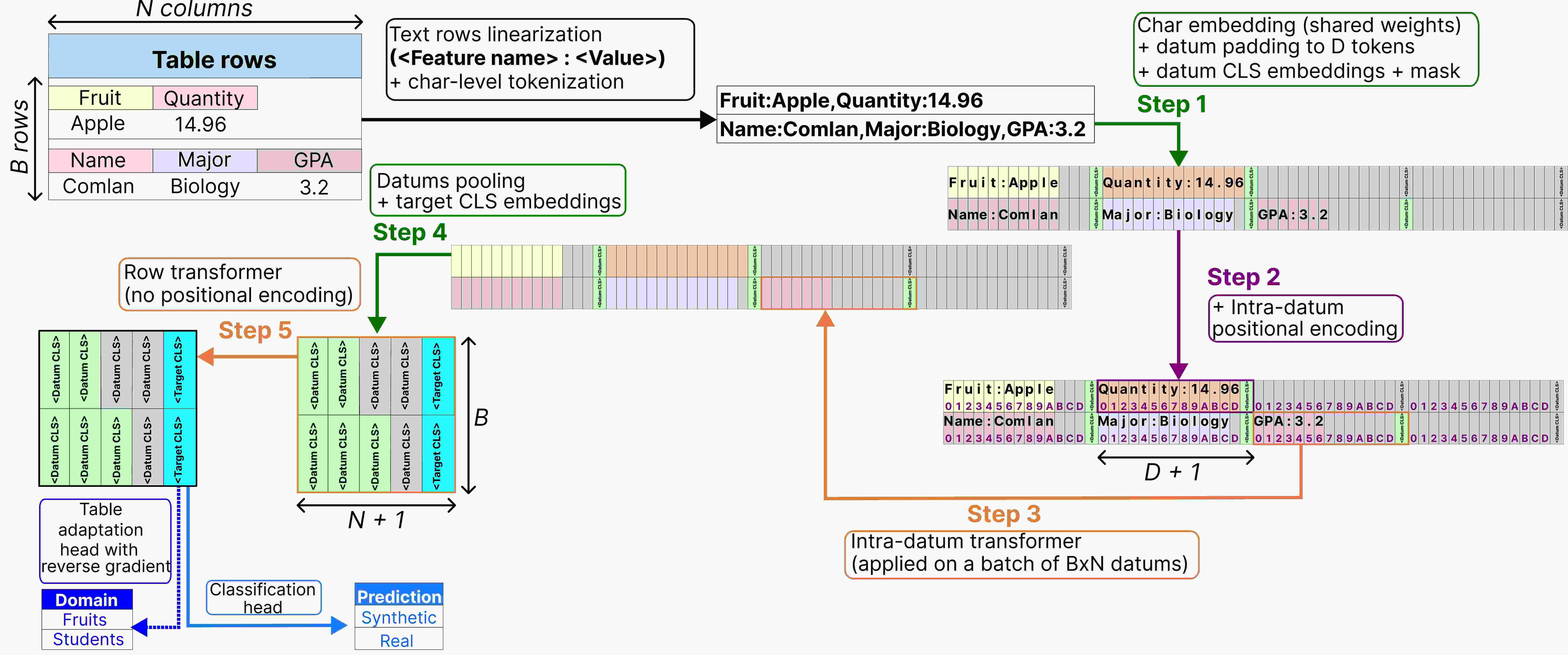}
    \caption{Datum-wise transformer pipeline with table adaptation head.}
    \label{fig:equivarch}
\end{figure*}

In the following, we describe our method for detecting synthetic tabular data. First, we describe the \textit{datum-wise} architecture, which is designed to generate effective representations for the synthetic data detection task.
We then describe the procedure used for \emph{table adaptation} to improve performance further.

\paragraph{Datum-Wise Transformer Architecture}

The proposed detector uses two transformers as its backbone: a \textit{datum transformer} and a \textit{row transformer}. The \textit{datum transformer} processes batches of text \textit{datums}, and the \textit{row transformer} works on a pooled \textit{datum} representation. The whole pipeline and architecture are described in Figure~\ref{fig:equivarch}. 

Each table row is converted into text (i.e. \texttt{datums}), which is the concatenation of \texttt{<column>:<value>} strings. The  \texttt{datums} are then tokenized at a character level. 
Technically, in the first step (Step 1 in Figure~\ref{fig:equivarch}), our model applies two levels of padding. \textit{Intra-datum} padding extends the length of each \textit{datum} to match the longest \texttt{<column>:<value>} string. Then, \textit{extra-datum} padding adds dummy \textit{datums} to handle varying numbers of \textit{datums} in each table of the training set. Each \textit{datum} is appended with a \textit{CLS} token, serving as a representation of the feature. In the following sections, we refer to these tokens as \textit{CLS-Datums}. 

A key architectural feature of our model is the restriction of positional encoding to individual \textit{datums} (Step 2 in Figure~\ref{fig:equivarch}). Processing the \textit{datums} independently as a first stage enables an independent ``featurization'', where the feature encodings are inferred directly from raw data. %
Operating at the \textit{datum} level, rather than on entire rows, also results in shorter input sequences for the \textit{datum transformer}, thereby reducing computational costs.
We avoid the reliance on column order induced by global positional encoding, which can cause problems when the detector is applied to tables with different column arrangements. We evaluate the effect of this particular feature in Section \ref{section:ablation_studies}. The positional encoding within the \textit{datum transformer} enables it to focus on column-related information without being conditioned on any specific column order. Without it, the transformer would not distinguish the positions of the characters and the entire row would be viewed as a bag of characters (where for instance $elbow:201.1$ and $below:1.012$ would be considered identical). While positional encoding matters within a \textit{datum}, it is important for tabular data, especially in synthetic data detection, to produce predictions that are invariant to column permutations.
At the end of \textit{Step 2}, each character is represented as a $192$-dimensional embedding, along with the added positional encoding.
This representation serves as input to the \textit{datum transformer} in \textit{Step 3}. From its output, we extract only the \textit{CLS-Datum} embeddings, which aggregate information from both the column name and value (Step 4). %
This datum pooling operation drastically reduces the input size for the subsequent steps. We expect the \textit{CLS-Datums} to provide sufficient information to effectively represent the features.
The \textit{CLS-Datums} are appended with an additional row-level \textit{CLS} token (192-dimensional, as for the \textit{CLS-Datums}) that will serve for our classification task. In the following paragraphs, we will refer to this token as \textit{CLS-Target}. The result of this operation is given as input to the \textit{row transformer} (\textit{Step 5}). %
This transformer does not incorporate any positional encoding as all the position-related information are already processed by the \textit{datum transformer}. We extract the \textit{CLS-Target} from the output and pass it to the classification head. Our detector is trained using a binary cross-entropy loss.

\paragraph{Table Adaptation}

To enhance our detector's performance ``in the wild'', we employ a domain generalization strategy, training on labelled data from multiple tables and using adversarial regularization to encourage invariance to table-specific artefacts. The model is evaluated on its ability to generalize to entirely unseen tables. In our context, we call it \textit{table adaptation}. %
Specifically, we employ the gradient reversal techniques from~\cite{ganin2015unsupervised,Saito_2018_ECCV} to minimize the classifier's reliance on table structures in its embeddings while emphasizing the values within the cells of the tables.
For example, some tables may exhibit characteristics that make them easily identifiable (e.g., all numeric versus mixed), encouraging the model to exploit these spurious signals rather than learning to distinguish real from synthetic rows more generally. In practice, we add a table classification head in our architecture from which the gradient reversal will be applied down to the representation learning layers. This table classification head predicts the name of the table and also utilizes the \textit{CLS-Target} produced by the \textit{row transformer} for its predictions. This is shown in the bottom left part of the Figure~\ref{fig:equivarch}. We still use a cross-entropy loss for optimization.

\section{Experimental Setup}
\label{section:experiments}

\subsection{Training Data}
\label{section:training_data}

\begin{table}[t!]
\small

    \centering
    
    \setlength{\tabcolsep}{1mm}
    
    \begin{tabular}{ccccc}\hline
        Table Domain & Name & Size & \#Num &  \#Cat\\\hline
        \multirow{4}{*}{\makecell{Societal \& \\ Demographic}} &Adult\footref{fn:dataset_link_openml} & 48842 & 6 & 9 \\
        &HELOC\footref{fn:dataset_link_kaggle} & 5229 & 23 & 1 \\
        &House 16H\footref{fn:dataset_link_openml} & 22784 & 17 & 0 \\
        &King\footref{fn:dataset_link_kaggle} & 21613 & 19 & 1 \\

        \midrule
        \multirow{5}{*}{\makecell{Consumer \& \\ Financial Behavior}} &Bank Marketing\footref{fn:dataset_link_openml} & 45211 & 7 & 10 \\
        &Churn Modelling\footref{fn:dataset_link_kaggle} & 4999 & 8 & 4 \\
        &Diamonds\footref{fn:dataset_link_openml} & 26970 & 7 & 3 \\
        &Black Friday\footref{fn:dataset_link_openml} & 166821 & 6 & 4 \\
        &Insurance\footref{fn:dataset_link_kaggle} & 1338 & 4 & 3 \\

        \midrule
        \multirow{4}{*}{\makecell{Science \& \\ Environment}} & Abalone\tablefootnote{\label{fn:dataset_link_openml}\url{https://www.openml.org}} & 4177 & 7 & 2\\
        &Cardio\tablefootnote{\label{fn:dataset_link_kaggle}\url{https://www.kaggle.com/datasets}} & 70000 & 11 & 1 \\
        &Higgs\footref{fn:dataset_link_openml} & 98050 & 28 & 1 \\
        &Bike Sharing\footref{fn:dataset_link_openml} & 17379 & 9 & 4 \\
        &MiniBooNE\footref{fn:dataset_link_openml} & 130064 & 50 & 1 \\\hline
    \end{tabular}
    \caption{Description of the tables considered in the experiments. We categorize the tables into three main domains (\textit{Social}, \textit{Finance}, and \textit{Science}) for the \textit{cross-domain table shift} evaluation. ``Size'' is the number of total instances in the table, ``\#Num'' and ``\#Cat'' refer respectively to the number of numerical and categorical attributes.}
    \label{table:datasets}
\end{table}

The tables were obtained from the OpenML repository and the Kaggle platform (see Appendix~\ref{appendix:table_links}). %
To make the detection task reasonably challenging, we employ state-of-the-art generators for creating synthetic data. Poorly trained or low-quality generators might produce data that is easily distinguishable from real data, hence, we prioritize using high-quality generators with carefully optimized hyperparameters for each table considered. The generators are TabDDPM~\cite{kotelnikov2023tabddpm}, TabSyn~\cite{zhang2023mixed}, TVAE, and CTGAN~\cite{CTGAN}. Hyperparameters were selected following the protocol presented in \cite{kindji2024hoodtabulardatageneration}. We employ the same tables as in the mentioned articles to leverage the available pretraining.

To train the detectors to distinguish between real and synthetic data, we construct each table as a balanced mix of both types of data as illustrated in Figure~\ref{fig:crosstable}. Specifically, for a given table with $n$ real rows, we add $n$ synthetic rows composed of $n/4$ rows from each of the four generators, resulting in a final table with $2n$ rows. 
This design maintains a balance between real and synthetic data by ensuring equal contribution from each generator within each table, thereby preventing any imbalance that could bias the detectors. Besides, it avoids introducing additional variables, such as varying real-to-synthetic ratios, that could interfere with the evaluation at this stage. However, it is important to note that the global balance across all tables is not enforced (e.g., one table may contain $2,000$ rows, while another may contain $20,000$). This variability reflects natural differences between tables while preserving local balance within each table.

\subsection{Baselines}
\label{section:baselines}

As discussed in  Section~\ref{section:related_work}, very few table encoders can be readily adapted to create a synthetic tabular data detector for real-world use. We view the \textit{Flat Text Transformer} (detailed below) as our closest baseline, given its alignment with our objectives. Nonetheless, we adapted other approaches to our goals to facilitate a more comprehensive comparison. %
Implementation and training details of our model and the considered baselines are provided in Appendix~\ref{appendix:hyperparameters}.

\subsubsection{Flat Text Transformer~\cite{KindjiIDA2025}} This model is explicitly designed for the cross-table setting and processes fully textual rows. It uses character-level tokenization with a global positional encoding across the entire row and employs a lightweight BERT-like transformer trained from scratch.

\subsubsection{TaBERT embedding~\cite{yin20acl}}

TaBERT can be reasonably adapted to our setup, due to its use of text-based row linearizations like \texttt{<column>|<type>|<value>}. Since it is designed to encode entire tables, we only considered its pretrained version setting the number of rows in each table to $1$ (\textit{K=1}~\footnote{\url{https://github.com/facebookresearch/TaBERT}}). This allows us to use it in our row-by-row classification task. \textit{TaBERT\textsubscript{base}} is initialized from \textit{BERT\textsubscript{base}}, which has $12$ heads and $12$ layers of attention.
Each row in our pool of tables is considered as a single table and encoded by \textit{TaBERT\textsubscript{base}}. The row context in natural language was generated by prompting \textit{GPT-4O-mini} to describe each table in our pool (see Table~\ref{table:datasets}). TaBERT\textsubscript{base} then processes a row formatted as a table, along with an associated context, as recommended by the authors. We retrieved each row's \textit{CLS} embedding and train a classification head. As the authors noted, their model can be viewed as an encoder for systems that require table embeddings as input.

\subsubsection{BART embedding and fine-tuned~\cite{lewis-etal-2020-bart}} 

BART is a general-purpose pretrained encoder, capable of directly processing our textualized row format for sequence classification. We evaluate both pretrained embeddings and a fine-tuned version of this model, using the same \textit{bart-base} checkpoint with 12 attention heads and 6 layers. In both cases, the BART tokenizer is used to ensure consistency with the model's input format. 
For the pretrained version, we deployed the model using the same procedure as for TaBERT. We generate embeddings for each row and extract the first token acting as a \textit{CLS} token and use it as input for the classification head. During training, only the weights of the classification head are updated.

Among the other pretrained models mentioned in Section~\ref{section:related_work}, we evaluated PORTAL using its original table-specific protocol, wherein a separate model was trained and deployed for each test table. While effective in this context, we did not include PORTAL as a \textit{cross-table shift} baseline due to the extensive refactoring and tuning required to generalize it across unseen tables.

\subsection{Detection Setups}
\label{s:detection_setup}

\paragraph{Cross-table Shift} In this setup, the model is deployed on unseen tables, as illustrated in Figure~\ref{fig:crosstable}. Note that our implementation involves a \textit{cross-table shift} between the train and validation sets, as well as between the train and test sets.

For our \textit{datum-wise} method, we evaluate two versions: one with table adaptation and one without. The experiments are conducted under a strict 3-fold cross-validation procedure. Performance of all detectors is reported using the ROC-AUC and accuracy metrics. %
ROC-AUC offers a threshold-independent measure of a detector's ability to distinguish real from synthetic data, essential in imbalanced or uncertain scenarios. Accuracy complements this by reporting performance at a fixed decision threshold of $0.50$.

\paragraph{Cross-domain Table Shift} 

We consider an additional setup with the same characteristics as the \textit{cross-table shift} but which involves tables from different domains between training and deployment. The domain distinctions used are presented in Table~\ref{table:datasets}. For simplicity, we refer to the table domains as \textit{Social}, \textit{Finance}, and \textit{Science}. To efficiently estimate performance without the additional overhead of cross-validation, we report metrics using bootstrapping~\cite{Efron1994bootstrap} and provide confidence intervals.

\section{Results}
\label{section:results}

We provide the experimental results for the \textit{cross-table shift} setup in Section~\ref{section:results_table_shift}, for the \textit{cross-domain table shift} one in Section~\ref{section:results_cross_domain}, and for a detailed analysis in Section~\ref{section:ablation_studies}. We discuss the limitations of our method in Section~\ref{section:limits}. The main results are reported in Table~\ref{tab:all_results} but additional results are reported in the text and detailed in the appendices.  

\subsection{Cross-table Shift}
\label{section:results_table_shift}

\begin{table}[t!]

\centering

\begin{tabular}{ccc}
\toprule
\multirow{2}{*}{Model} & \multicolumn{2}{c}{Metrics} \\
        \cline{2-3}
                                                    & AUC & Accuracy \\ 
            \midrule

          BART-embd & $0.50 \pm 0.00 $ & $0.50 \pm 0.00$ \\
          \midrule
          BART fine-tuned & $0.52 \pm 0.03$ & $0.52 \pm 0.02$ \\

        \cline{1-3}
          TaBERT-embd & $0.51 \pm 0.00$ & $0.50 \pm 0.00$ \\
        
        \cline{1-3}
        
         Flat Text  & $0.60 \pm 0.07$ & $0.52 \pm 0.01$ \\
         
        \cline{1-3}
         Datum-wise  & $0.67 \pm 0.05$ & $0.59 \pm 0.08$ \\

        \cline{1-3}
         Datum-wise + TA  & \bm{$0.69 \pm 0.04$} & \bm{$0.66 \pm 0.05$} \\
\bottomrule
\end{tabular}
\caption{AUC and Accuracy performance (mean $\pm$ standard deviation) for transformer detectors. Our proposed \textit{Datum-wise} method is evaluated with and without table adaptation (TA). ``BART-embd'' and ``TaBERT-embd'' refer respectively to the embeddings produced by BART and TaBERT.}
\label{tab:all_results}

\end{table}

The results from our experiments on the \textit{cross-table shift} setup are reported in Table~\ref{tab:all_results}. We provide the performance for the considered baselines and our \textit{datum-wise} method evaluated with and without table adaptation using table names as domains.
Our method consistently outperforms all baselines across all metrics, achieving an average AUC of $0.67$ and accuracy of $0.59$, establishing state-of-the-art results for the \textit{cross-table shift} detection setup. In comparison, the best baseline, \textit{Flat Text}, tailored for this task, reaches an AUC of $0.60$ and accuracy of $0.52$. We observed consistent improvements in all three folds, though statistical significance testing is not reliable at this scale.

As for BART and TaBERT's embeddings (respectively \textit{BART-embd} and \textit{TaBERT-embd}), we observe that they achieve performance close to random, with an AUC of $0.50$ for BART and $0.51$ for TaBERT. Both models obtain an accuracy of $0.50$. 
Analyzing the training logs reveals a notable improvement in performance on the training set, with an average AUC of $0.63$ for TaBERT's embedding and $0.59$ for BART. However, both models struggle to generalize to the validation and test sets. 

These observations highlight a broader limitation when repurposing pretrained models designed for language understanding or reasoning over structured text for tasks involving synthetic data detection in tabular domains. Both TaBERT and BART were adapted with care to fit our row-by-row classification setting, leveraging configurations (e.g., TaBERT with \textit{K=1}) and processing pipelines that remain as faithful as possible to their architectural expectations. Nonetheless, the fundamental shift in task (from textual reasoning to real versus synthetic row classification across heterogeneous table structures) presents challenges that these models were not originally optimized to address. In particular, the lack of full-table context and the absence of inter-row dependencies reduce their effectiveness in this setting.

The fine-tuned BART model achieves an average AUC of $0.52$, slightly outperforming the \textit{BART-embd} baseline, which scores $0.50$. %
To investigate these limited results, we analyzed embeddings extracted from the decoder output before the classification head on the first fold. A T-SNE~\cite{JMLR:v9:vandermaaten08a} visualization (Appendix~\ref{appendix:embeddings_tsne}) showed that the model mainly relies on table-specific characteristics, with clear separations between tables. This was confirmed quantitatively by training an XGBoost classifier on the $768$-dimensional embeddings to predict table names, achieving $0.99$ accuracy.

\subsubsection{Table Adaptation Strategy} 

In this configuration, a parameter \textit{lambda} regulates the intensity of gradients propagated from the table classification head. This parameter is gradually increased from $0$ to $1$ over the course of training. Initially, the model undergoes a few iterations to learn the primary target classification task, i.e., distinguishing between real and synthetic data, before being exposed to negative gradients.
This gradual introduction avoids early suppression of table-specific features, ensuring the model has time to learn informative patterns for the primary task.

Our initial experiments with the original scheduling approach proposed by~\cite{ganin2015unsupervised} showed it to be overly aggressive for our setup, often leading to early stopping after just a few epochs. To mitigate this issue, we implemented a smoother, cosine-based lambda schedule, which ultimately yielded the best performance, as demonstrated in Table~\ref{tab:all_results}. With this adjusted table adaptation strategy, the accuracy improved from $0.59$ to $0.66$, and the AUC increased from $0.67$ to $0.69$. The simultaneous improvement in both metrics suggests that the model initially relied on table-related features, among other factors. This influence was effectively mitigated through the adapted training strategy. We further analyze the impact of the table adaptation strategy in a detailed analysis presented in Section \ref{section:ablation_studies}.

\subsection{Cross-domain Table Shift}
\label{section:results_cross_domain}
 
\begin{table}[t]
    \centering

\setlength{\tabcolsep}{1mm}

    \begin{tabular}{l|ccc|c}
    \toprule
    Train \textbackslash Test & Social & Finance & Science \\
    \midrule

    Social   & - & $0.49 \pm 0.00$ & $0.48 \pm 0.00$  \\
    Finance   & $0.48 \pm 0.00$  & - &  $0.50 \pm 0.00$  \\
    Science    & $0.48 \pm 0.00$ & $0.51 \pm 0.00$ & - \\
    
    \midrule

    SocialAFN  & -  & $0.49 \pm 0.00$ & $ \bm{0.52 \pm 0.00}$ \\
    FinanceAFN  &  $0.50 \pm 0.00$ & - & $\bm{0.55 \pm 0.00}$  \\
    ScienceAFN  & $ 0.50 \pm 0.00$ & $0.51 \pm 0.00$ & - \\

    \bottomrule
    \end{tabular}
    \caption{Cross-domain AUC for the \textit{datum-wise} classifier trained on one domain and tested on others. This setting combines both a domain shift and a cross-table shift. We report the average performance over $500$ bootstrapped tests along with the 95\% confidence intervals. ``AFN'' refers to anonymized features and noise (perturbed rows).}
    \label{table:cross_domain_exploration_experiments}
    \end{table}

 According to Table~\ref{tab:all_results}, the \textit{datum-wise} model combined with table adaptation is able to generalize well across different table structures.
 However, in order to evaluate the ability of the model to generalize across both domains and structures, we partitioned the tables into three domains: \textit{Social}, \textit{Finance}, and \textit{Science} (See Table~\ref{table:datasets}). We then trained the \textit{datum-wise} classifier on the tables from one domain and evaluated its ability to detect synthetic content on tables from other domains. This \emph{cross-domain table shift} setting is extremely difficult as it combines both a \textit{cross-domain shift} and a \textit{cross-table shift}.
  As expected, it does not work (AUC around $1/2$ is equivalent to a random decision). The results of this experiment are reported at the top of Table~\ref{table:cross_domain_exploration_experiments}. They suggest that cross-table generalization is only possible among semantically similar tables.
A limitation of this experiment, however, is that the domain-specific models were only trained on subsets of the dataset we used for the main \textit{cross-table shift} experiment of Table~\ref{tab:all_results}.
This reduced coverage likely contributes to the poor out-of-domain generalization.

To investigate this problem, we analyzed the embeddings by extracting the \textit{CLS-Target} and visualizing them using T-SNE plots (see Appendix~\ref{appendix:embeddings_tsne}). This revealed a strong reliance on table characteristics when predicting the outcome (real or synthetic). In response, we explored various strategies aimed at mitigating this behavior and improving out-of-domain generalization.
We anonymized all table features by replacing original column names with generic feature indices (e.g.,  \texttt{feature\_<i>:<value>}) to prevent the model from relying on specific column names. Additionally, we simulated realistic noise in synthetic data and replaced 20\% of the rows in each synthetic table with noisy versions. %
Categorical values in synthetic tables were replaced with either a generic placeholder, a random scrambled string drawn from the column's character set, or a shuffled permutation of a randomly chosen existing category (e.g., replacing ``apple'' with a shuffled ``orange'', like ``aegnor''). Additional details are in Appendix~~\ref{appendix:data_perturbation_protocol}. Perturbations replace synthetic rows only in the source domain, also maintaining a balance between real and synthetic data.

Each strategy was evaluated individually and in combination. The combination of anonymized features (AF) and added noise (N) in the synthetic data yielded the best performance for the detection task, both in-domain and out-of-domain. The results (with suffix AFN) are reported at the bottom of Table~\ref{table:cross_domain_exploration_experiments}. 
We observe a slight out-of-domain performance improvement when adapting to the \textit{Science} domain; for instance, adapting from \textit{Finance} to \textit{Science} yields an AUC increase from $0.50$ to $0.55$. Though modest, this gain indicates that the strategy may enhance cross-domain adaptation.

\begin{table*}[]
\centering
\begin{tabular}{|c|l|l|cccc|}
\hline
\multirow{2}{*}{Setup}  & \multirow{2}{*}{Model}     & \multirow{2}{*}{Training Data} & \multicolumn{4}{c|}{Evaluation Data}                                                                                             \\ \cline{4-7} 
& &                                & \multicolumn{1}{c|}{Train} & \multicolumn{1}{c|}{Train (Perm.)} & \multicolumn{1}{|c|}{Test} & Test (Perm.) \\ \hline
\multirow{3}{*}{\makecell{Cross-table\\ Shift}}  & \multirow{2}{*}{Flat Text} & Original                          & \multicolumn{1}{c|}{$\bm{0.74 \pm 0.02}$}             & \multicolumn{1}{c|}{$0.67 \pm 0.03$}             & \multicolumn{1}{|c|}{$0.60 \pm 0.07$}           &   \multicolumn{1}{c|}{$0.60 \pm 0.08$}            \\ \cline{3-7} 
 & & Dynamic Perm.                          & \multicolumn{1}{c|}{$0.67 \pm 0.03$}             & \multicolumn{1}{c|}{$0.66 \pm 0.04$}             & \multicolumn{1}{|c|}{$0.60 \pm 0.08$}           &    \multicolumn{1}{c|}{$0.61 \pm 0.06$}          \\  \cline{2-7}
& Datum-wise            & Original                          & \multicolumn{1}{c|}{${0.72 \pm 0.02}$}             & \multicolumn{1}{c|}{$\bm{0.72 \pm 0.02}$}             & \multicolumn{1}{|c|}{$\bm{0.69 \pm 0.04}$}           &    \multicolumn{1}{c|}{$\bm{0.69 \pm 0.04}$}           \\ 
\midrule
\midrule
\multirow{3}{*}{\makecell{Single Table\\ \small (Black Friday)}}  & \multirow{2}{*}{Flat Text} & Orig.                          & \multicolumn{1}{c|}{$0.65 \pm 0.00$}             & \multicolumn{1}{c|}{$0.51 \pm 0.00$}             & \multicolumn{1}{|c|}{$0.64 \pm 0.01$}           & $0.52 \pm 0.01$             \\ \cline{3-7} 

& & Dynamic Perm.                          & \multicolumn{1}{c|}{$0.51 \pm 0.00$}             & \multicolumn{1}{c|}{$0.55 \pm 0.00$}             & \multicolumn{1}{|c|}{$0.51 \pm 0.01$}           &  
                         $0.54 \pm 0.01$           \\ \cline{2-7}
& Datum-wise           & Orig.                          & \multicolumn{1}{c|}{$\bm{0.66 \pm 0.00}$}             & \multicolumn{1}{c|}{$\bm{0.66 \pm 0.00}$}             & \multicolumn{1}{|c|}{$\bm{0.66 \pm 0.01}$}           & $\bm{0.66 \pm 0.01}$             
\\

\bottomrule
\end{tabular}
\caption{AUC and standard deviation results for a text transformer with global positional encoding (\textit{Flat Text}) and our column-permutation invariant model (\textit{Datum-wise}). At the top we use the \textit{cross-table shift} setting as in Table~\ref{tab:all_results}, at the bottom we perform the same experiment on a \textit{single-table}: Black Friday. We compare the evaluation on unchanged datasets and column-permuted datasets (``Perm.'').
``Dynamic Perm.'' applies a random column shuffling during training. We use $500$ bootstrapped tests and $95\%$ confidence intervals.}
\label{table:col_permutation_study}
\end{table*}

\subsection{Detailed Analysis}
\label{section:ablation_studies}

We conduct a detailed analysis to evaluate the impact of individual components within the datum-wise method.

\subsubsection{Impact of Table Adaptation}

To evaluate the impact of table adaptation, we extracted and visualized the \textit{CLS-Target} embeddings from the \textit{row transformer} just before the classification and adaptation heads (Appendix~\ref{appendix:embeddings_tsne}).

We computed the average pairwise cosine distance between L2-normalized table centroids in our 192-dimensional embedding space. After adaptation, embeddings became more clustered, reducing inter-table distances from $3.30 \times 10^{-5}$ to $6.17 \times 10^{-6}$ (a $81.3\%$ relative decrease) indicating a more unified representation. Although absolute distances are small due to normalization, this relative reduction indicates a positive effect of the table adaptation strategy. To further assess inter-table separability, we trained an XGBoost classifier to predict the table identity from the row embeddings. The classifier's accuracy dropped from $0.99$ before adaptation to $0.89$ after adaptation, confirming that the embeddings encode less table-specific information and thus exhibit reduced inter-table separability.

\subsubsection{Permutation Invariance}

A widely used method, especially in image processing, to enforce invariance of a predictive model to a group of transformations, such as symmetry or rotation, is to augment the training data by applying these transformations randomly to the training instances.
In image processing, this \textit{data augmentation} procedure is considered essential to improve the generalization ability of the models, but it requires a longer training phase and it does not provide strong guarantees, especially if the transformation space is large. A better option is to use neural architectures that are explicitly designed to be invariant or equivariant to these transformations \cite{cohen2016group, xu20232}. These architectures provide a better generalization, especially when facing distributional shift \cite{pmlr-v139-elesedy21a}.
Regarding tabular data, there is no spatial ordering relationship between features; it is hence natural to seek the invariance by permutation of the columns. In \cite{borisovlanguage23}, the authors propose to train their models on random permutations of the columns, but this strategy is not guaranteed to cover all possible permutations.

Our goal here is to assess the gain of our \textit{datum-wise} model that was explicitly designed for column permutation invariance (referred to as \textit{Datum-wise}) against an equivalent text transformer with global positional encoding (referred to as \textit{Flat Text}). For the \textit{Flat Text} model, we consider two training strategies: one without permutation of the columns (referred to as ``Original'') and one with permutation of the columns (referred to as ``Dynamic Perm.'').
We consider two tasks: synthetic tabular data detection under \textit{cross-table shift} as in Table~\ref{tab:all_results} and a \textit{single table} setting (here, the \textit{Black Friday} table).
We evaluate the obtained models on both permuted and non-permuted sets.
These results, reported in Table~\ref{table:col_permutation_study}, confirm that our \textit{datum-wise} model generalizes better than a classical text transformer with global positional encoding.

In the sub-column ``Train (Perm.)'' we provide the results obtained when \textit{evaluating} the models on a permuted version of its training data. These results, compared to the sub-column ``Train (Orig.)'', confirm that permutation alone (without changing the cell values in the tables) strongly degrades the \textit{Flat Text} model performance but does not impact the \textit{Datum-wise} model.
The dynamic column permutation strategy (``Dynamic Perm.'') seems not to improve significantly the test performance of the \textit{Flat Text} model, which remains behind the \textit{Datum-wise} model.
A larger-scale experiment could reveal a slight improvement of the \textit{Flat Text} model with dynamic column permutation, but the differences we obtain between ``Test'' and  ``Test (Perm.)'' are not significant for the \textit{cross-table shift} setting. This could be explained by the fact that the test column names in our test tables do not appear in the training sets or are too different from the ones encountered in the training tables. 

We also performed the same experiment on a \textit{single table} (the \textit{Black Friday} table) to see if the dynamic column permutation strategy would improve the generalization ability of the \textit{Flat Text} model when facing the same features names, but it seems on contrary 
to overfit strongly the position of the columns and their permutations. As expected, the \textit{Datum-wise} model remains insensitive to these perturbations. Full training dynamics and analysis on permutation proportions are provided in Appendix~\ref{appendix:permutation_invariance}.

\subsection{Limitations}
\label{section:limits}

Our approach focuses solely on row-by-row detection. However, we acknowledge that some statistical properties, especially for time-series or sequential data, may become apparent only when considering multiple rows together. %

Moreover, while our tables are diverse, they may not capture the full spectrum of real-world data, especially from specialized or proprietary domains. Evaluating on more application-specific tables and synthetic generators would better test the generality and robustness of our findings. In addition, the cross-domain generalization remains a key challenge and points to the need for more data and more advanced adaptation techniques.

Another limitation of our approach is its potential difficulty in handling column name ambiguity, where the same column name or abbreviation can refer to different concepts across tables, especially when the tables are from different domains. An interesting solution could be to equip our model with a table context/domain encoder to resolve these ambiguities.
Additionally, our model is designed to be invariant to column permutations, but we do not systematically study other perturbation types, such as missing values, adversarial modifications, all of which may impact detector reliability in practice. This work also lacks a theoretical analysis of the effects of positional encoding, particularly regarding its influence on the model's effectiveness in detecting synthetic data and its impact on permutation invariance. 

Finally, although our \textit{datum-wise} Transformer is more lightweight than standard BERT-based encoders, we do not provide a detailed evaluation of computational efficiency or scalability for large-scale or real-time deployments. Intermediate pooling strategies, such as extracting additional tokens beyond \textit{CLS-Datums}, may also enhance representation capacity, but remain unexplored in this work.

\section{Conclusion}
\label{section:conclusion}
In this study, we address the underexplored challenge of detecting synthetic tabular data in real-world scenarios, where models must generalize to unseen table structures. We introduced a novel \textit{datum-wise} Transformer architecture that operates on character-level embeddings and employs local positional encoding at the column level. Our method significantly outperforms the sole existing baseline for this task and other purposefully designed competitors. 
Through a thorough evaluation under both cross-table and cross-domain protocols, we demonstrated that generalizing across table structures presents a greater challenge than domain shift alone. Further, we showed that incorporating a lightweight table adaptation strategy can significantly enhance performance. 
Our results provide the first compelling evidence that robust detection of synthetic tabular data in real-world conditions is not only possible but can be effectively achieved with tailored architectures and targeted adaptations. This architecture opens up several promising avenues for future research, such as supporting pretraining-finetuning pipelines for tabular prediction tasks, using objectives like Masked Language Modeling (MLM) from TaBERT, or employing few-shot learning strategies like STUNT.

\section{Acknowledgement}

This work was granted access to the HPC resources of IDRIS under the allocations  AD011014381R1, AD011015150R1, and AD011012220R2 made by GENCI.

\bibliography{biblio}

\appendix

\renewcommand{\thesection}{\arabic{section}}

\section{Reproducibility Details}
\label{appendix:reproducibility_details}

In this section, we provide key details to ensure experimental reproducibility, including the sources of the tables (Appendix~\ref{appendix:table_links}), the hyperparameters and training procedures for the detectors (Appendix~\ref{appendix:hyperparameters}), and the hardware specifications (Appendix~\ref{appendix:hardware}).

\subsection{Source and References of Tables}
\label{appendix:table_links}

The tables used in this study are publicly available and widely used in the machine learning literature. Below, we provide appropriate citations and references for each table to ensure proper attribution and reproducibility.

\begin{itemize}
    \item \textbf{Abalone}~\cite{nash1994population}
    \item \textbf{Adult}~\cite{adult_2}
    \item \textbf{Bank Marketing}~\cite{bank_marketing_222}
    \item \textbf{Black Friday}~\cite{black_friday}
    \item \textbf{Bike Sharing}~\cite{bike_sharing_275}
    \item \textbf{Cardio}~\cite{cardio} 
    \item \textbf{Churn Modelling}~\cite{churn}
    \item \textbf{Diamonds}~\cite{diamonds}
    \item \textbf{HELOC}~\cite{fico2018heloc}
    \item \textbf{Higgs}~\cite{higgs_280}
    \item \textbf{House 16h}~\cite{house16h}
    \item \textbf{Insurance}~\cite{insurance_kaggle}
    \item \textbf{King}~\cite{kingcounty2014_2015}
    \item \textbf{MiniBooNE}~\cite{miniboone_particle_identification_199}
\end{itemize}

\subsection{Computing Infrastructure}
\label{appendix:hardware}

Experiments were conducted on machines equipped with a Tesla V100-SXM2 (32 GB) provided by IDRIS. Additional runs were performed on a workstation with two NVIDIA GeForce RTX 4090 GPUs (24GB each), running under CUDA 12.4 and NVIDIA driver version 550.144.03. The system had an AMD Ryzen 9 CPU, 128GB of RAM, and operated under Ubuntu 22.04 LTS. All models were implemented in Python using PyTorch 2.4.1 and Hugging Face Transformers 4.45.2. Additional key libraries included NumPy 2.1.2, scikit-learn 1.5.2, and pandas 2.2.3.

\subsection{Hyperparameters and Training Procedures}
\label{appendix:hyperparameters}

\begin{table}[h]
    \centering
    \begin{tabular}{ll}
        \toprule
        \textbf{Hyperparameter} & \textbf{Value} \\
        \midrule
        Batch Size            & 64 \\
        Number of Layers      & 3 \\
        Number of Heads       & 6 \\
        Embedding Dimension   & 192 \\
        Dropout               & 0.2 \\
        Learning Rate         & $1 \times 10^{-5}$ \\
        \bottomrule
    \end{tabular}
    \caption[Hyperparameters used for the \textit{Flat Text} and \textit{Datum-wise} models.]{Hyperparameters used for the \textit{Flat Text} and \textit{Datum-wise} models in all experiments. For the \textit{Datum-wise} model, both the \textit{datum-transformer} and the \textit{row-transformer} share the same architecture.}
    \label{table:transformers_hyperparameters}
\end{table}

We use the same set of hyperparameters for both the \textit{Flat Text Transformer} and the \textit{Datum-wise Transformer}, as presented in Table~\ref{table:transformers_hyperparameters}. Training is performed with the Adam~\cite{adamKingma2014} optimizer. The classification head consists of a batch normalization layer, followed by a linear layer and a Sigmoid activation function. In the \textit{datum-wise} model, the additional domain head (used for table classification) has the same structure but uses a Softmax activation instead of Sigmoid.

For the pretrained encoder baselines, we use BART and TaBERT. Both models produce 768-dimensional embeddings, which are subsequently fed into \emph{their own} classification heads; these heads share the same architecture as those used in the transformer-based models but are not shared across baselines. Each baseline is trained with the Adam optimizer and a learning rate of $5 \times 10^{-5}$, corresponding to the default setting in Hugging Face's \textit{TrainingArguments}\footnote{\url{https://huggingface.co}}.
The models are trained for 10 epochs, with early stopping criteria applied if there is no improvement on the validation AUC over three successive epochs. This number of epochs has been sufficient, as all models stopped training before reaching the maximum limit, with each epoch taking approximately one and a half hours. For fine-tuning BART, we trained for $5$ epochs and selected the best model based on the validation set. This setting was sufficient, as the model consistently began to overfit beyond this point.

All neural network-based detectors are trained with a cross-entropy loss.

\section{Embeddings}
\label{appendix:embeddings_tsne}

During our experiments on the \textit{cross-table shift} and \textit{cross-domain table shift} settings, t-SNE visualizations were used to gain insights into how the detectors separate tables in their embedding spaces. These visualizations are provided here for reference.

\begin{figure*}[ht]
    \centering

    \subfigure[Before Table Adaptation]
    {%
        \includegraphics[width=0.7\linewidth]{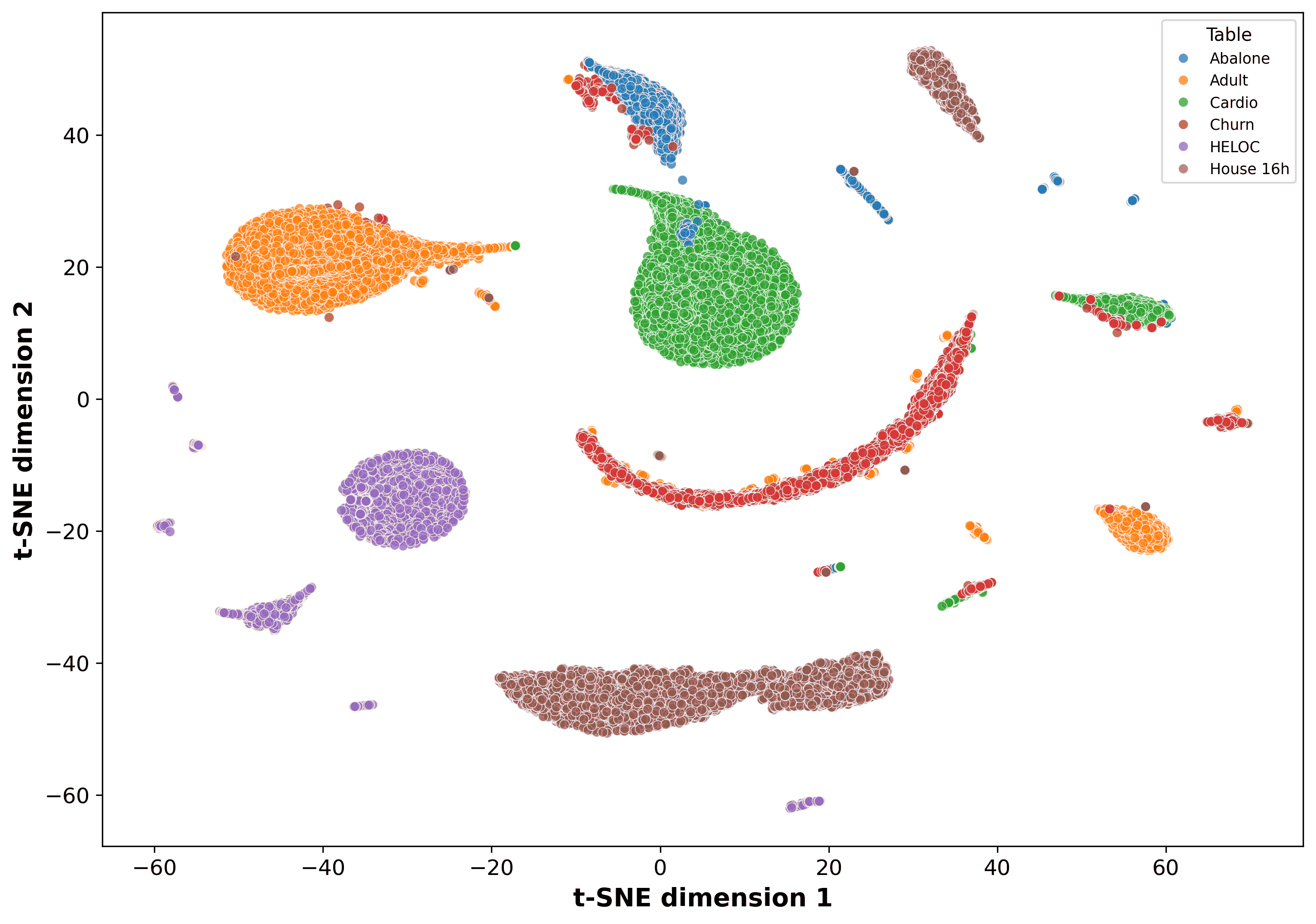}
        \label{fig:datum_embd_before_da}
    }
    \subfigure[After Table Adaptation]
    {%
        \includegraphics[width=0.7\linewidth]{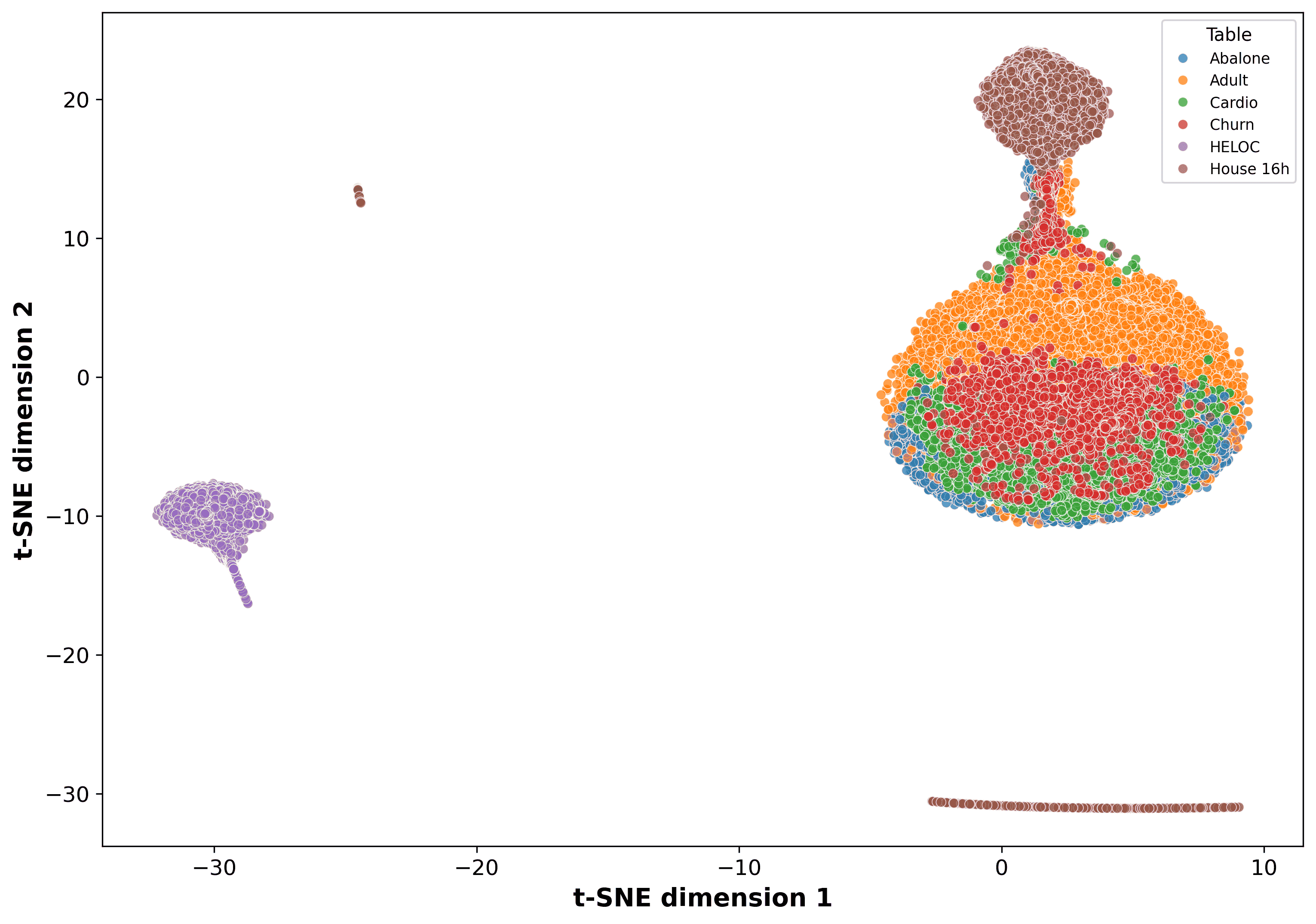}
        \label{fig:datum_embd_after_da}
    }
    \caption[t-SNE projection of row embeddings before and after table adaptation, colored by table.]{t-SNE projection of row embeddings colored by table. The embeddings are extracted from the trained \textit{datum-wise} model before and after the table adaptation strategy considering table names as domains.} 
    \label{figure:datum_embds_before_after_da}
\end{figure*}

\begin{figure*}[]
    \centering
    \includegraphics[width=0.7\textwidth]{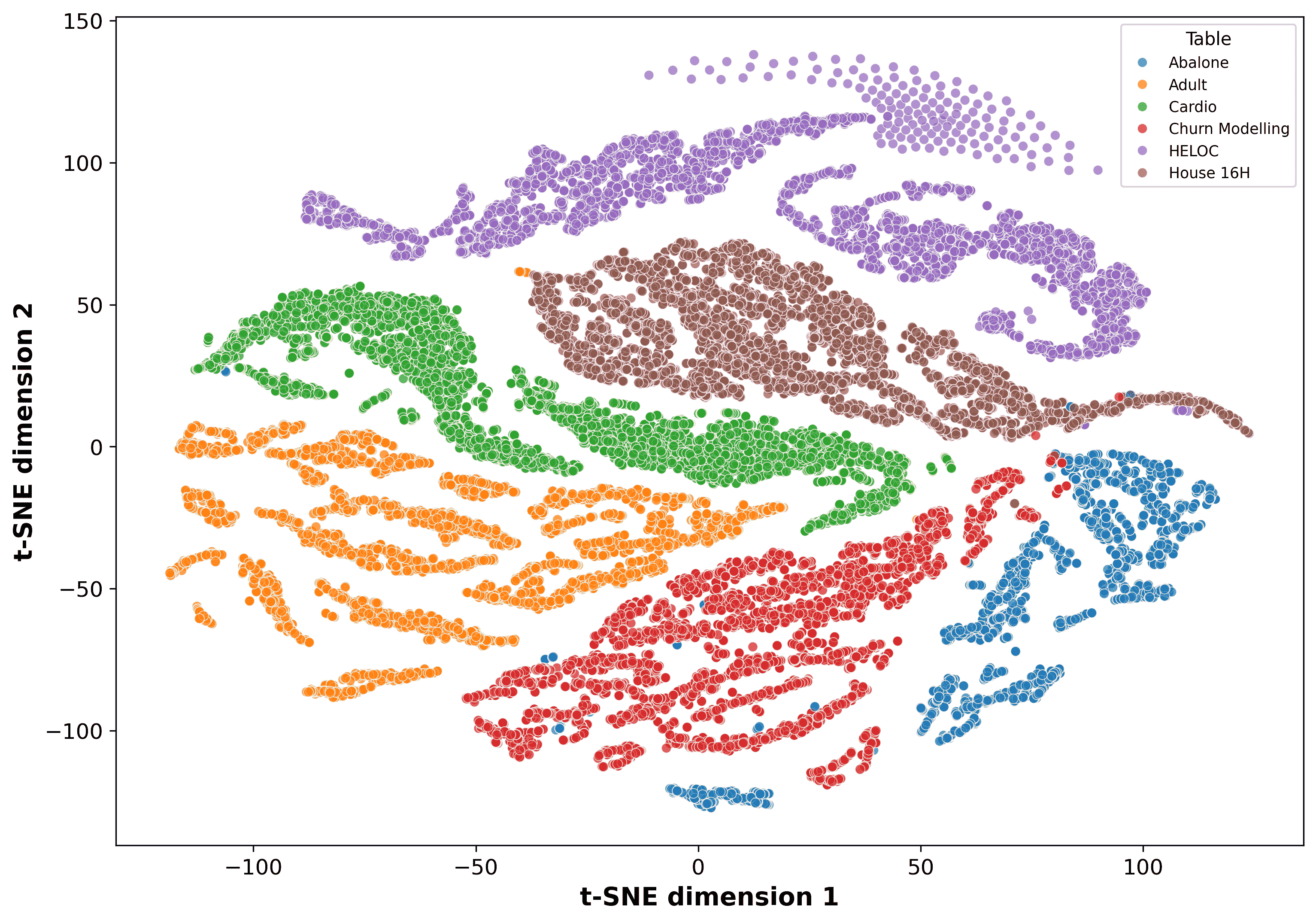}
    \caption[t-SNE projection of row embeddings for the fine-tuned BART baseline, colored by table.]{t-SNE projection of row embeddings colored by table for the fine-tuned BART baseline on the first fold of the \textit{cross-table shift} setting.}
    \label{figure:bart_row_embd_test_tables}
\end{figure*}

\begin{figure*}[]
    \centering
    \includegraphics[width=0.7\textwidth]{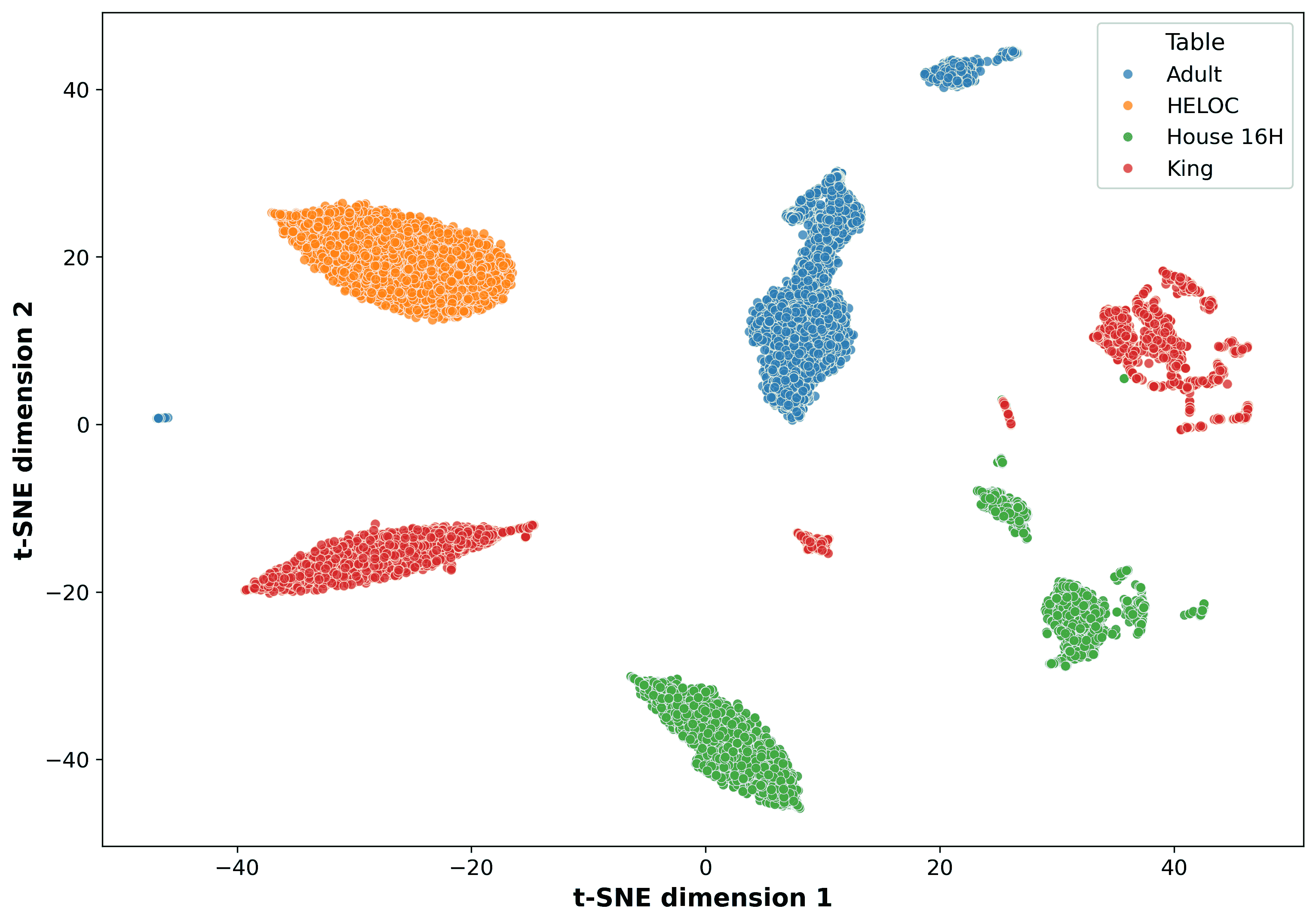}
    \caption[t-SNE projection of row embeddings for the \textit{datum-wise} model in the \textit{Social} table domain.]{t-SNE projection of row embeddings colored by table for the \textit{datum-wise} model in the \textit{Social} domain under the \textit{cross-domain table shift} setting.}
    \label{figure:social_tables_tsne}
\end{figure*}

\begin{figure*}[]
    \centering
    \includegraphics[width=0.7\textwidth]{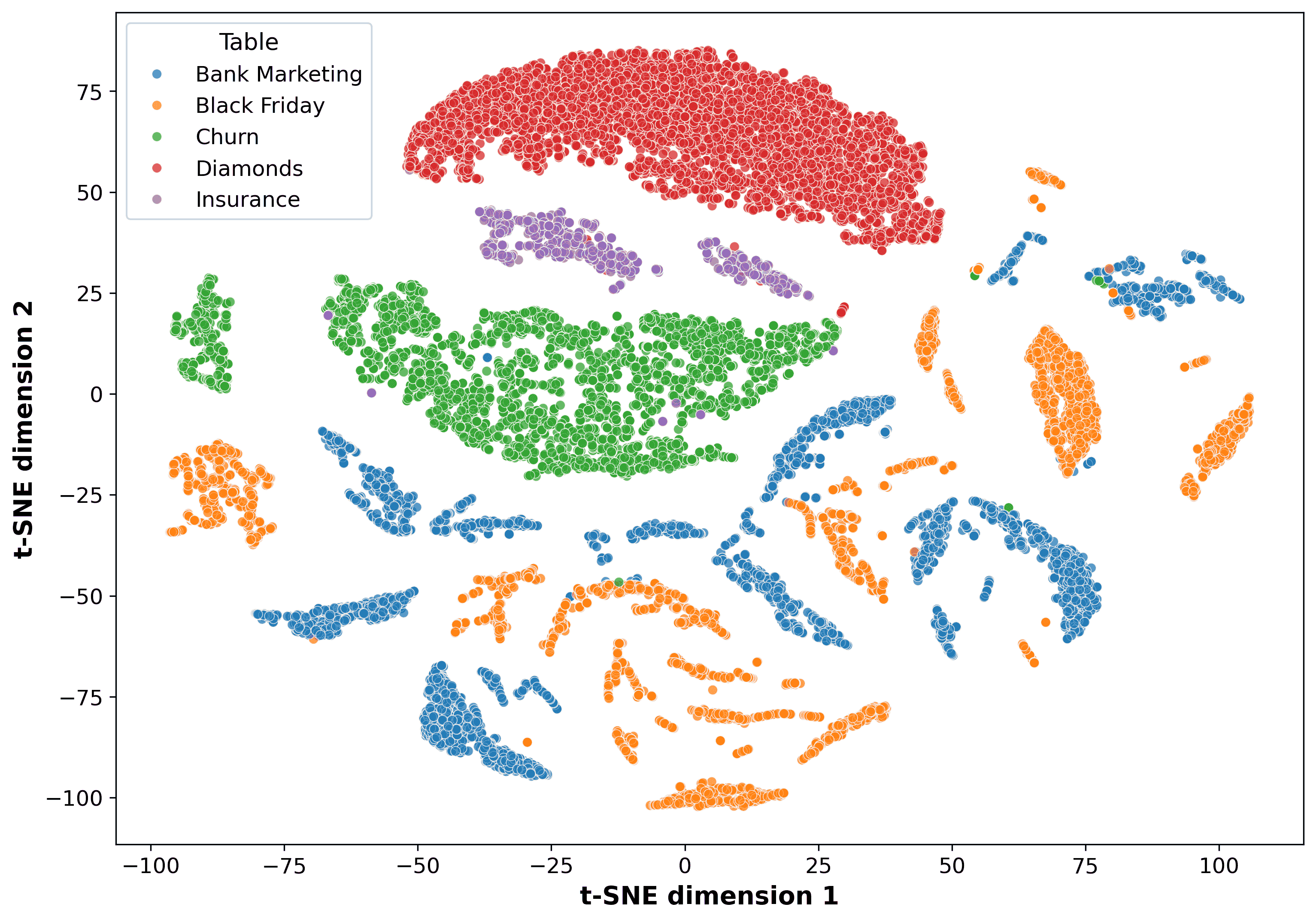}
    \caption[t-SNE projection of row embeddings for the \textit{datum-wise} model in the \textit{Finance} table domain.]{t-SNE projection of row embeddings colored by table for the \textit{datum-wise} model in the \textit{Finance} domain under the \textit{cross-domain table shift} setting.}
    \label{figure:finance_tables_tsne}
\end{figure*}

\begin{figure*}[]
    \centering
    \includegraphics[width=0.7\textwidth]{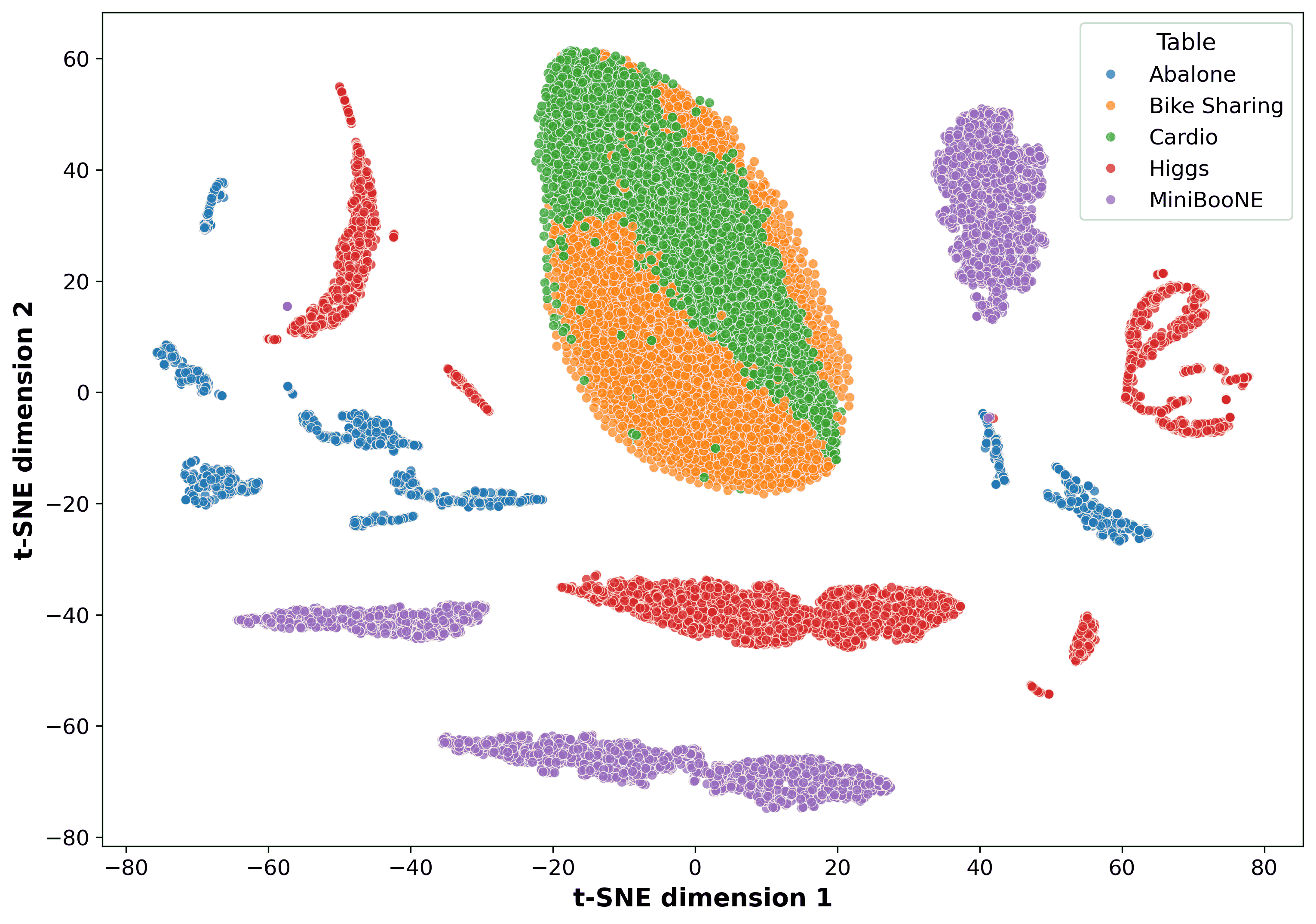}
    \caption[t-SNE projection of row embeddings for the \textit{datum-wise} model in the \textit{Science} table domain.]{t-SNE projection of row embeddings colored by table for the \textit{datum-wise} model in the \textit{Science} domain under the \textit{cross-domain table shift} setting.}
    \label{figure:science_tables_tsne}
\end{figure*}

\begin{itemize}
    \item \textbf{Figure~\ref{figure:datum_embds_before_after_da}:} The \textit{datum-wise} model before and after the table adaptation strategy, on validation and test tables for the first fold split in the \textit{cross-table shift} setting.
    \item \textbf{Figure~\ref{figure:bart_row_embd_test_tables}:} The fine-tuned BART baseline on validation and test tables for the first fold split in the \textit{cross-table shift} setting.
    \item \textbf{Figures~\ref{figure:social_tables_tsne}, \ref{figure:finance_tables_tsne}, and \ref{figure:science_tables_tsne}:} Embeddings of the \textit{datum-wise} model under the \textit{cross-domain table shift} setting, corresponding to models trained on the \textit{Social}, \textit{Finance}, and \textit{Science} domains, respectively. These figures illustrate how the model separates its training data (in-domain).
\end{itemize}

Figure~\ref{figure:datum_embds_before_after_da} (\textit{cross-table shift} setting) illustrates how the table adaptation strategy reduces table separability in the embedding space, promoting more homogeneous representations. Complementary visualizations in Figures~\ref{figure:social_tables_tsne}, \ref{figure:finance_tables_tsne}, and \ref{figure:science_tables_tsne} (from the \textit{cross-domain table shift} setting) reveal more distinct table-specific clusters, indicating the detector's reliance on table-related features, among other factors. The models in this setting were trained on smaller subsets of the data used in the main \textit{cross-table shift} experiments, which likely contributed to the observed drop in generalization. These visual patterns helped motivate the strategies that led to modest improvements in out-of-domain performance, as discussed in Section~\ref{section:results_cross_domain}. Overall, the findings underscore the importance of both table adaptation and sufficient data coverage when addressing synthetic tabular data detection across structurally diverse tables. 

\section{Data Perturbation}
\label{appendix:data_perturbation_protocol}

As described in the main text (Section~\ref{section:results_cross_domain}), after observing limited out-of-domain generalization in the \textit{cross-domain table shift} setting, we implemented several strategies to improve performance. One such approach involved applying a data perturbation protocol, where 20\% of the synthetic tables were replaced with noisy versions of the rows. Additional details about the perturbation protocol are provided in this section.

For categorical features, consider a column called ``\textit{Fruit}'' with possible values ``\textit{apple}'' and ``\textit{orange}'', and suppose the vocabulary extracted from the synthetic table is $V = \{a, p, l, e, o, r, n, g\}$. The perturbation consists of replacing a random subset of values with one of three types of noise to simulate realistic perturbations:

\begin{itemize}
    \item the string ``???'', indicating a missing or unknown category (e.g., ``\textit{apple}''~$\rightarrow$~``???'');
    \item a scrambled string composed of randomly selected characters from the vocabulary \( V \), with length varying between 3 and 10 characters (e.g., ``\textit{prn}'', ``\textit{rngoppe}'');
    \item a shuffled version of an existing category string from the same feature (e.g., replacing ``\textit{apple}'' with ``\textit{pleap}'', or ``\textit{orange}'' with ``\textit{aegnor}'').
\end{itemize}

These perturbations are motivated by the need to preserve the original characters (since our model leverages character-level embeddings) while introducing noise that maintains a plausible categorical pattern.

For numerical features, noise is introduced by replacing selected values with random numbers uniformly sampled within the original column's minimum and maximum range, maintaining the overall data distribution while adding variability. It is important to note that the noisy rows come in replacement of the initial synthetic rows and not in addition, as it would introduce a table-level imbalance we wanted to avoid for the original experiment. Also, it is worth noting that no noise was added in the target domains.

\section{Permutation Invariance}
\label{appendix:permutation_invariance}

\begin{figure*}[]
    \centering
    \includegraphics[width=0.7\textwidth]{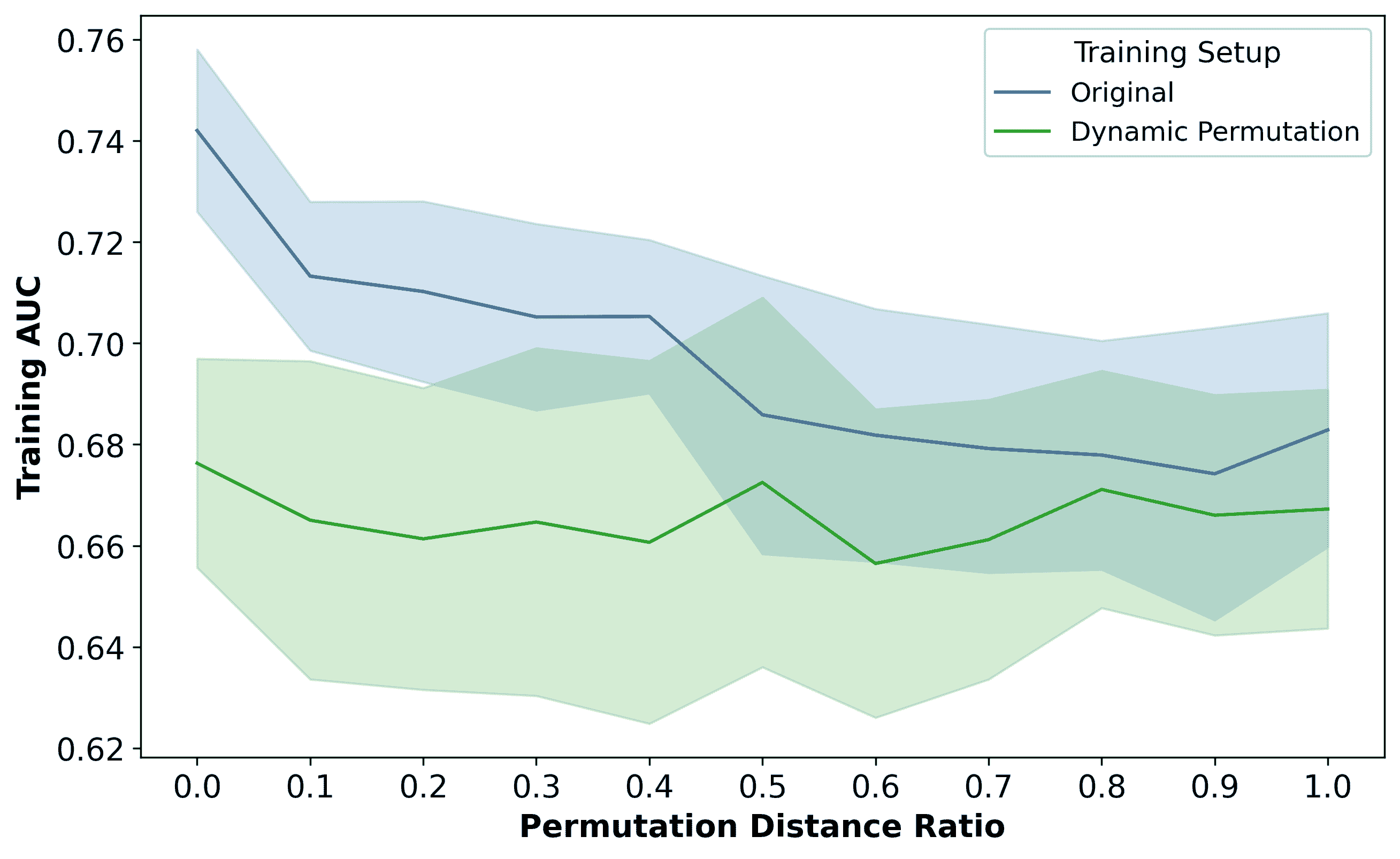}
    \caption[Robustness to column permutations of the \textit{Flat Text} Transformer.]{\textit{Flat Text} Transformer: AUC performance as a function of column permutation distance at \textbf{train} time, across different training setups \textbf{for the cross-table shift}. 
The \textit{Original} model is trained with fixed column order; the \textit{Dynamic Permutation} model is trained with a different random permutation per sample.}
\label{fig:permutation_distance_evaluation_tab_shift}
\end{figure*}

\begin{figure*}[ht]
    \centering

    \subfigure[Train]
    {%
        \includegraphics[width=0.7\linewidth]{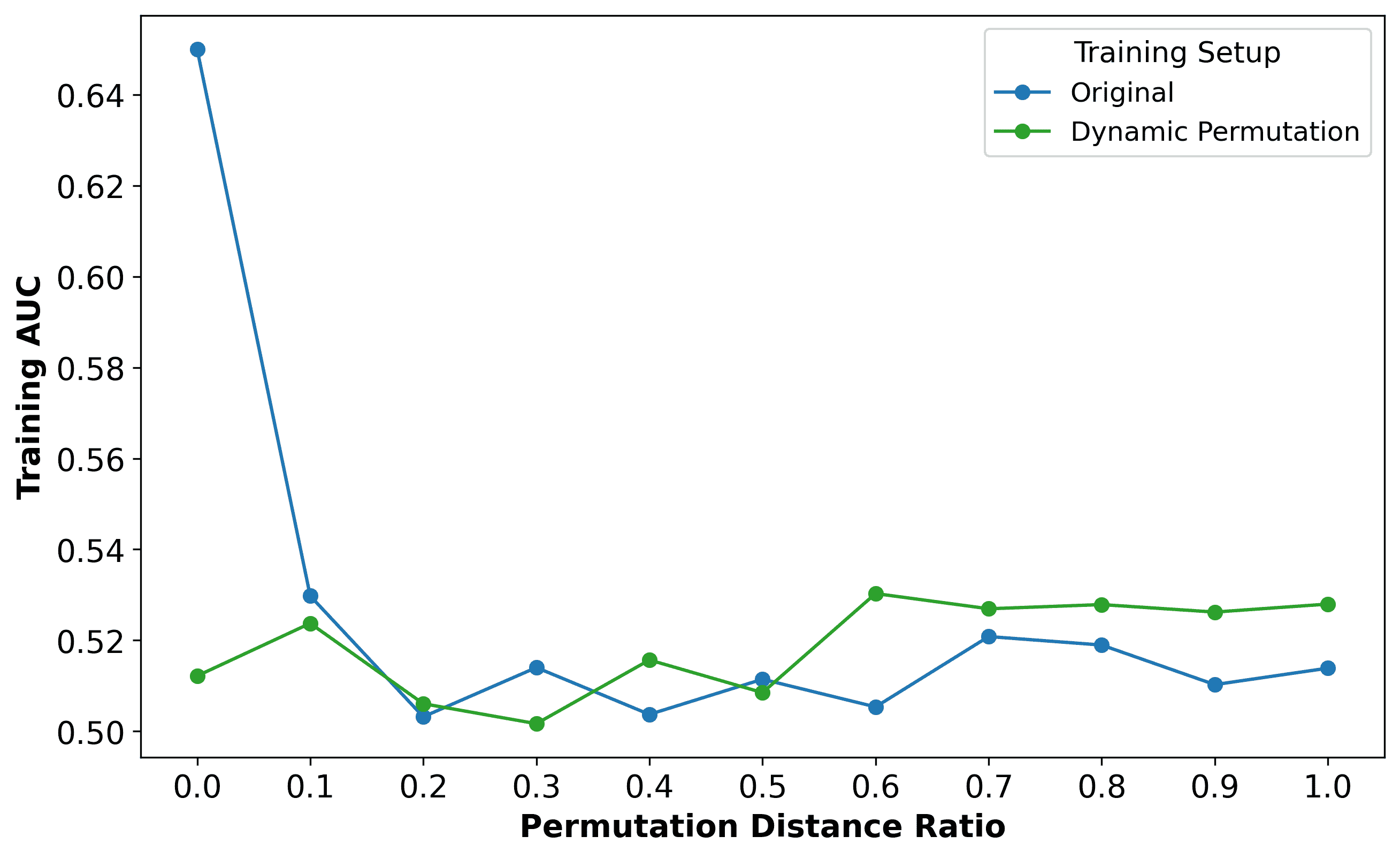}
        \label{fig:perm_distance_train_single_table}
    }
    \subfigure[Test]
    {%
        \includegraphics[width=0.7\linewidth]{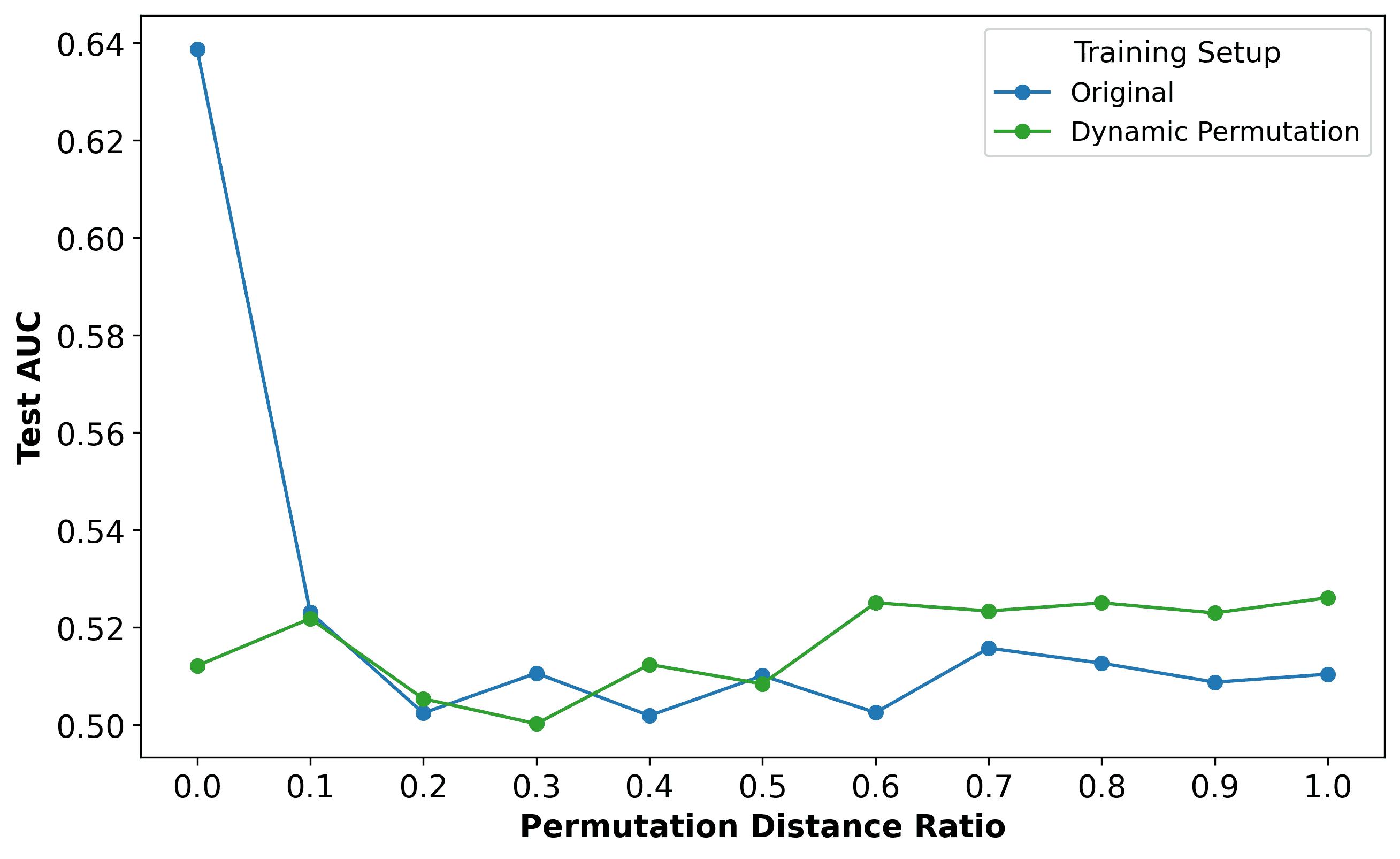}
        \label{fig:perm_distance_test_single_table}
    }
    \caption[\textit{Flat Text} baseline: AUC performance as a function of column permutation distance.]{\textit{Flat Text} baseline: AUC performance as a function of column permutation distance at \textbf{train} (\ref{fig:perm_distance_train_single_table}) and \textbf{test} time (\ref{fig:perm_distance_test_single_table}), across different training setups \textbf{for the \textit{single table}} setting. 
The \textit{Original} model is trained with fixed column order; the \textit{Dynamic Permutation} model is trained with a different random permutation per sample.} 
    \label{figure:perm_distance_single_table}
\end{figure*}

As demonstrated in Section~\ref{section:ablation_studies}, our \textit{datum-wise} method exhibits strong robustness across permuted versions of tables. To further understand the role of permutation sensitivity, we investigate the performance variability of the \textit{Flat Text Transformer} baseline under various perturbations. This analysis provides additional insight into how such permutations affect models that are not explicitly designed to be permutation-invariant. %
We evaluate performance variations based on \textit{permutation distance}, defined as the proportion of permuted columns in a table, ranging from $0.0$ (no permutation) to $1.0$ (full permutation). The results are presented in Figure~\ref{fig:permutation_distance_evaluation_tab_shift} for the \textit{cross-table shift}. The results for the \textit{single table} at both training and test time are also provided in Figure~\ref{figure:perm_distance_single_table} for reference. As a reminder, the \textit{single table} approach (considering the \textit{Black Friday} table here) follows the ``\textit{same-table detection}'' setting as described in Section~\ref{section:intro}.

We notice similar trends for the \textit{cross-table shift} and the  \textit{single table} setup at both training and test time. %
As described in the main text, the \textit{cross-table shift} scenario involves test tables that are completely separate from those in the training set. Because these test tables are new and unseen during training, the results obtained by permuting their columns are less meaningful.

As shown in Figures~\ref{fig:permutation_distance_evaluation_tab_shift} and~\ref{figure:perm_distance_single_table}, when trained on the fixed \textit{original} column order, the \textit{Flat Text Transformer} delivers declining performance as permutation distance increases, indicating a strong reliance on positional dependencies, confirming earlier conclusions from the main text. In contrast, when trained with dynamic column permutations, it maintains more stable performance, confirming reduced sensitivity to column order.

\end{document}